\documentclass{article}
\usepackage{spconf,amsmath,graphicx}
\usepackage{helvet}
\usepackage{courier}
\usepackage{amsmath}
\usepackage{amssymb}
\usepackage{subfigure}
\usepackage{subfigure}
\usepackage{algpseudocode}
\usepackage{algorithm}
\usepackage[american]{babel}
\usepackage{microtype}
\usepackage{bm}
\usepackage{graphicx}
\usepackage{url}
\usepackage{multirow}
\usepackage{amsfonts}
\usepackage{booktabs}       
\algdef{nSxnE}{Begin}{End}{}{}

\title{Adversarial Examples Detection beyond Image Space}
\name{\normalsize {Kejiang Chen$^{\star}$\ Yuefeng Chen$^{\dagger}$\ Hang Zhou$^{\star}$ \ Chuan Qin$^{\star}$\ Xiaofeng Mao$^{\dagger}$\ Weiming Zhang$^{\star}$\ Nenghai Yu$^{\star}$\thanks{This work was supported in part by the Natural Science Foundation of China under Grant U1636201, and by Anhui Initiative in Quantum Information Technologies under Grant AHY150400.
Contact mail: zhangwm@ustc.edu.cn}}}
\address{$^{\star}$ University of Science and Technology of China \\
	$^{\dagger}$ Alibaba Group}

\begin{document}
\ninept
\maketitle
\begin{abstract}
	Deep neural networks have been proved that they are vulnerable to adversarial examples, which are generated by adding human-imperceptible perturbations to images. To defend these adversarial examples, various detection based methods have been proposed. However, most of them perform poorly on detecting adversarial examples with extremely slight perturbations. By exploring these adversarial examples, we find that there exists compliance between perturbations and prediction confidence, which guides us to detect few-perturbation attacks from the aspect of prediction confidence.
	To detect both few-perturbation attacks and large-perturbation attacks, we propose a method beyond image space by a two-stream architecture, in which the image stream focuses on the pixel artifacts and the gradient stream copes with the confidence artifacts. The experimental results show that the proposed method outperforms the existing methods under oblivious attacks and is verified effective to defend omniscient attacks as well.
  \end{abstract}

  \section{Introduction}

  Deep neural networks have been very successful in recognizing visual objects, and state-of-the-art neural networks even perform better than humans on large-scale image classification tasks\cite{DBLP:conf/icml/TanL19}. However, their robustness has raised concerns, and recently researches show that they are fragile to adversarial-based perturbations\cite{DBLP:conf/iclr/BhattadCLLF20,DBLP:conf/ccs/CoMML19}. These adversarial examples are threatening if neural networks are utilized in crucial real applications, such as autonomous driving and identity recognition. 
  
  To solve this, plenty of works have been proposed to defend adversarial examples in DNNs, and can be roughly categorized into 1) \emph{defense} that focuses on making the underlying model robust to adversarial examples, and 2) \emph{detection} that attempts to distinguish adversarial example from innocent inputs \cite{huang2019model}. Most \emph{defense} methods\cite{DBLP:conf/sp/PapernotM0JS16,DBLP:conf/iclr/MadryMSTV18,DBLP:conf/iccv/MummadiBM19} modify the target models, and the expensive retraining process makes them impractical for massive data classification. Generally, the accuracy will decrease, which is not acceptable for big tasks, such as malicious image detection. The \emph{detection} can be deployed in bypass without affecting the original task. Additionally, it can also be used in conjunction with robust defense.
  
  There are many detection methods proposed recently from different aspects, including prediction logits~\cite{FS,liang2018detecting}, pixel artifacts~\cite{DBLP:journals/corr/FeinmanCSG17,liu2019detection} and the layer consistency~\cite{LID,MAD}. SRM~\cite{SRM} owns the-state-of-art performance, which detects the artifacts from the aspect of steganalysis. However, SRM depends heavily on the artifacts. Recently, novel attacks, such as Decoupled Direction and Norm (DDN)~\cite{DDN} and Elastic-net Attacks to DNNs (EAD)~\cite{EAD}, deceive the classification model with few or exiguous perturbations, and the experiments show that these adversarial examples cannot be detected by SRM effectively. 
  
  \begin{figure}[t]
	\centering 
		\includegraphics[width=3in]{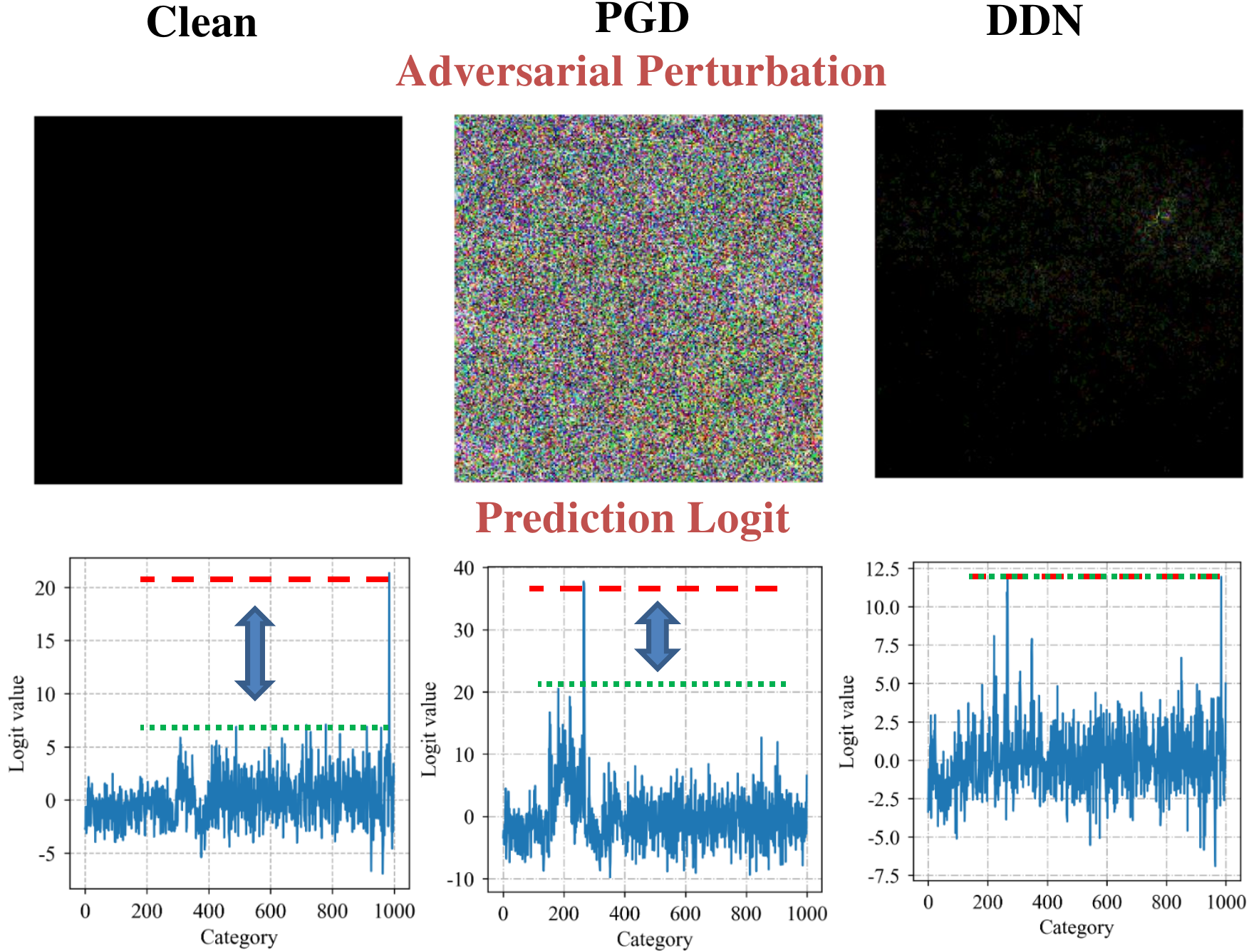}
		\vspace{-1em}
	\caption{The comparison between the clean image and adversarial images in terms of perturbation (zoomed 30 times), prediction logits and confidence gradient. The confidence of slight perturbation adversarial example (DDN) is low, and that of adversarial examples with many and large perturbations (PGD) is high, indicating that there exists compliance between perturbation and prediction confidence, which guides us to detect adversarial examples from pixel artifacts and gradient artifacts.} 
	\label{compare} 
	\vspace{-1em}
  \end{figure}

  In this paper, we first analyze the prediction logits of clean examples and adversarial examples, and find that there exists compliance between perturbation and prediction confidence. Generally, the prediction confidence can be defined as the advantage of the rank-1 predicted logit to the rank-2 predicted logit. As shown in Figure \ref{compare}, for few-perturbation attack (DDN), the prediction confidence is low. For large perturbation attack (PGD), prediction confidence is high. In conclusion, the stronger the perturbation, the higher of the prediction confidence. The phenomenon indicates that the prediction confidence can be used for detecting few-perturbation attacks.  
  
Inspired by LID and MAD, we further propose confidence gradient
  to gather more discriminative information from the classification model. The confidence loss is defined as the cross-entropy between predicted logit and its one-hot version, representing prediction confidence. Afterwards, the confidence gradient is computed by back-propagation, which includes the information of both prediction confidence and classification model.
  
  For detecting few-perturbation attacks as well as large-perturbation attacks, we propose a novel adversarial example detection through exploiting both pixel artifacts and confidence artifacts (abbreviated as PACA). The method is under a two-stream framework, where the image stream is used to capture pixel artifacts and the gradient stream is used to catch gradient artifact.
  
  

  We apply our method to detect various attacking methods including $\ell_1, \ell_2$ and $\ell_\infty$ constraint on the widely used ImageNet and Caltech-256 datasets under different threat models including oblivious adversaries and omniscient adversaries, where the former adversaries only deceive the classification model and the latter adversaries know both the classification model as well as the detection model and try to deceive both.
  
  The results demonstrate that compared to the baseline, the proposed method improves the detection accuracy against adversarial attacks under oblivious adversaries in most cases. Besides, we demonstrate that the omniscient adversaries have to craft adversarial examples with larger noises to successfully mislead the classification equipped with our detection. 
  
	  
  


  \section{Related Work}
  
  There are many detection methods proposed recently from different aspects. ~\cite{DBLP:journals/corr/FeinmanCSG17,liu2019detection} detects adversarial examples by exploiting the image artifacts. Feature Squeezing (FS)~\cite{FS} processes the input image and discriminates according to the change of the prediction. Local Intrinsic Dimensionality (LID)~\cite{LID} and Mahalanobis Adversarial Detection (MAD)~\cite{MAD} detects adversarial examples based on the consistency within the model. It has been pointed out \cite{liu2019detection} that Spatial Rich Model (SRM)~\cite{SRM} owns the-state-of-art performance, which detects the artifacts from the aspect of steganalysis. SRM has superior performance than FS, LID, and MAD when detecting well-known attacks. However, we implement the detection task on the newly proposed methods with few perturbations, such as DDN and EAD, and find that they cannot be detected by SRM effectively, motivating us to design a novel method to detect these imperceptibly adversarial examples.

  \section{Methodology}
  In this section, we first analyze the properties of these examples, which guide us to the new method. 
  \subsection{Analysis}

\begin{figure}[t]
	  \centering 
		  \includegraphics[width=2.5in]{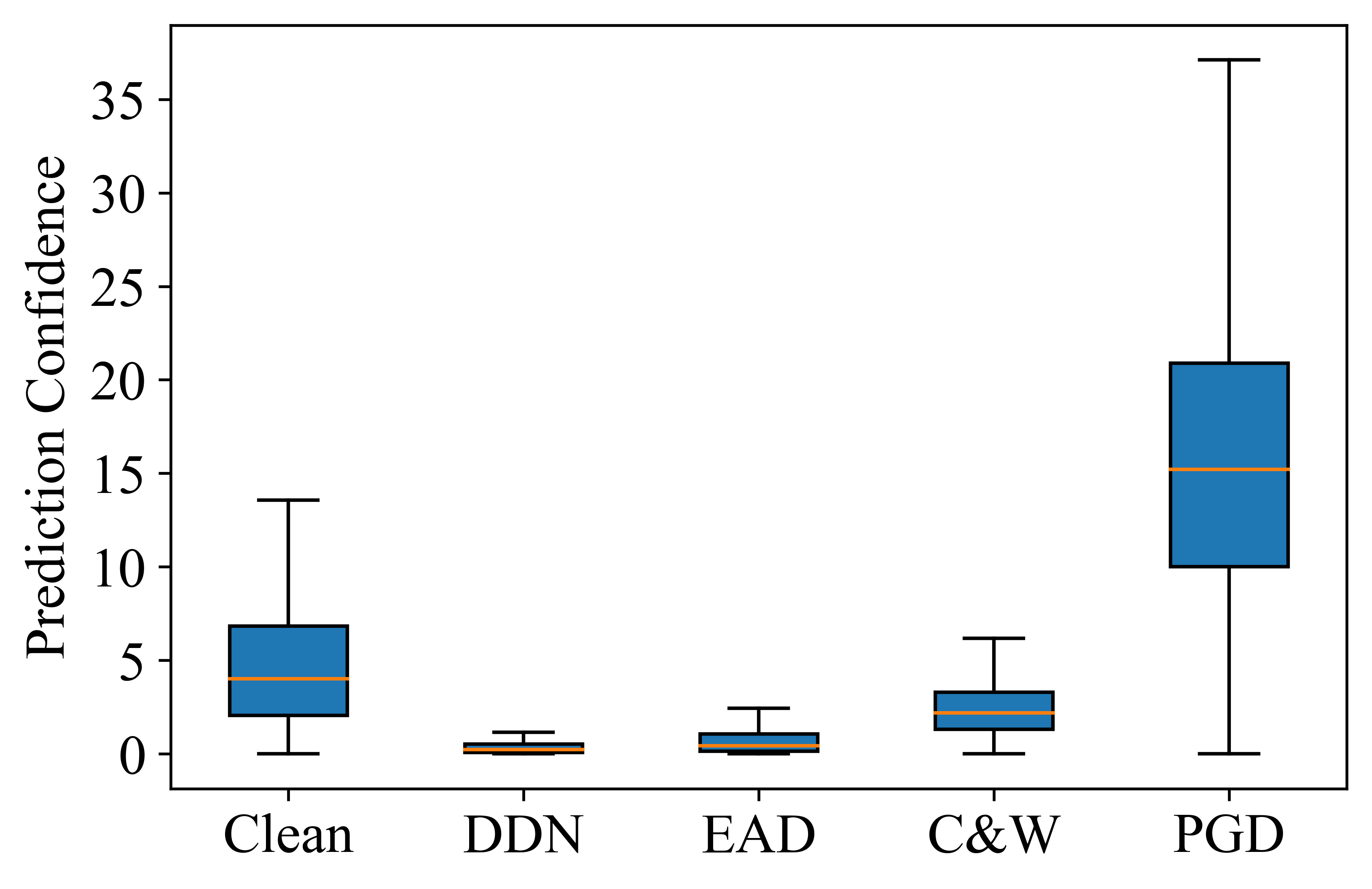}
		  \vspace{-1em}
	  \caption{The distribution of the prediction confidence of clean images and adversarial images.} 
	  \label{prediction_confidence} 
	  \vspace{-1em}
  \end{figure}

 Figure \ref{prediction_confidence} shows the distribution of prediction confidence of 500 clean images and different types of adversarial images. It is oblivious that the prediction confidence is discriminative between clean images and adversarial images crafted by few-perturbations attacks (DDN, EAD). This phenomenon implies us to detect few-perturbations attacks (DDN, EAD) from the perspective of confidence artifacts. For large-perturbation attacks, we can detect them from pixel artifacts.


  

  
  \begin{figure}[t]
	  \centering 
		  \includegraphics[width=3.5in]{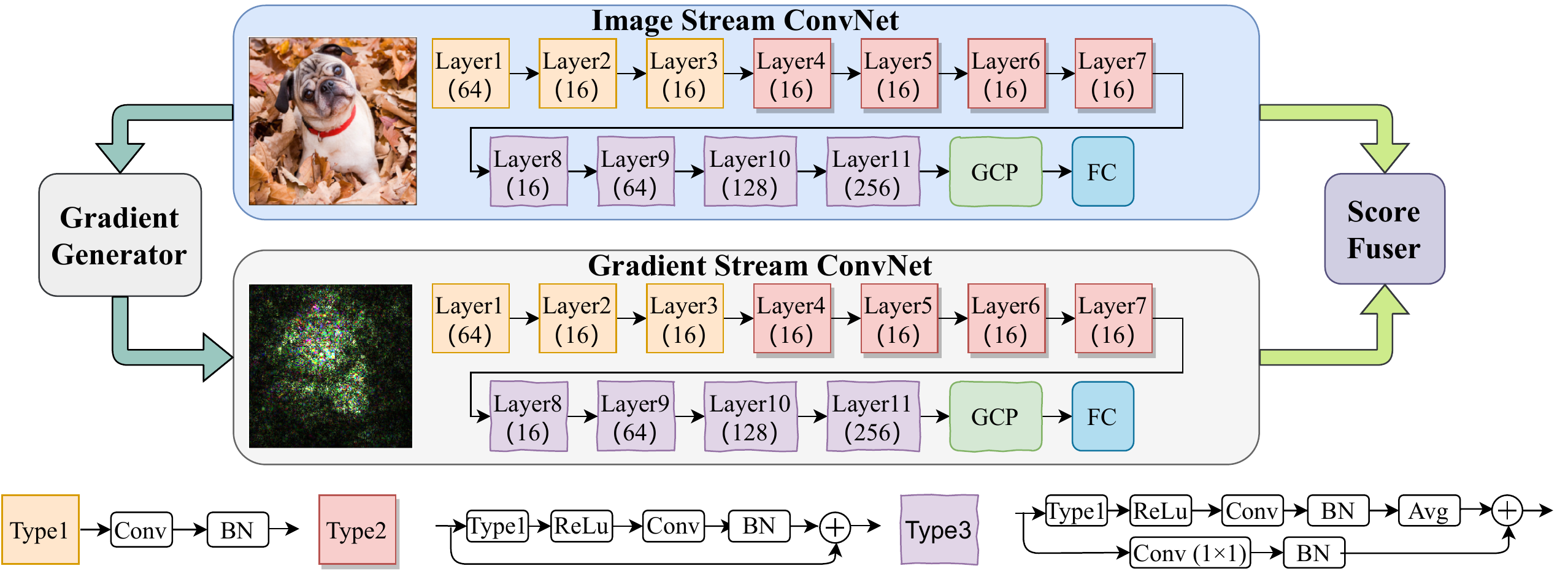}
		  \\\vspace{-1em}
	  \caption{The two-stream convolutional neural network for adversarial example detection. There are three different types of layers, shown in different shapes, and their architectures are defined at the bottom. The kernel size of the convolution in Layer 1 is $5\times5$ to obtain large reception field and others are $3\times3$ without specific instruction. The number in parentheses under the text ``Layer'' denotes the number of kernels. BN, GCP and FC represent batch normalization, global covariance pooling and fully connected layer, respectively.} 
	  \label{twostream} 
	  \vspace{-1em}
  \end{figure}


  \subsection{Pixel Artifacts and Confidence Artifacts (PACA)}
  
  To detect both large-perturbation attacks as well as few-perturbation attack, we propose a novel method to exploit both pixel artifacts and confidence artifacts by a two-stream architecture, named PACA.
  
  The PACA consists of a gradient generator, two identical sub-networks with different inputs, and a score fuser. Given an image, the gradient generator will generate its gradient, which reflects the information of both prediction confidence and classification model. Then the image and the gradient are fed to two sub-networks to get the immediate scores. Finally, the score fuser mixes the immediate scores, and output the final result. The detail of every part will be explained in the following subsections.

  \subsubsection{Gradient Generator}
  Drawing lessons from LID and MDA that the information of the model does help detection, we are about to gather more information from the classification model with the prediction confidence. 
    At first, we design a loss function, named confidence loss:
    \begin{equation}
  	  L = - \sum \nolimits _ { i } ^ { n } t _ { i } \log \left( y _ { i } \right)
    \end{equation}
    where $y _ { i } = \frac {  e ^ { z _ { i } } } { \sum _ { j } ^ { n} e ^ { z _ { j } } }$, $n$ is the class number, and $\bm{t}$ is the one-hot vector of the predicted logits $\bm{z}$.
	It should be noticed that one-hot vector $\bm{t}$ is the predicted label of the input image, not the true label. 
	Small confidence loss corresponds to the high confidence of the classification of the input image. For fully employing the information of the classification model, we compute the gradient of image by back-propagating confidence loss to the image. In implementation, the absolute value of gradient is fed into the sub-network.

  \subsubsection{Subnetwork}
  The image $\bm{x}$ and the gradient $|\bm{g}|$ are then fed to the sub-network. As for the image stream, the task of the backbone is to classify an input image as a clean or an adversarial image:
  \begin{equation}
	  \bm{y}=\begin{cases}
		  \bm{x},&		\text{clean}\\
		  \bm{x}+\boldsymbol{\delta},&		\text{adversarial}\\
	  \end{cases}
  \end{equation}
  where $\bm{x}$ is the clean image, and $\boldsymbol{\delta}$ denotes the adversarial perturbation. It should be noticed that the perturbation $\boldsymbol{\delta}$ is quite small compared to $\bm{x}$. Thus common neural architectures may not perform well on discrimination for they possibly diminishing the perturbation signal, i.e. average pooling will suppress noise-like perturbation signals by averaging adjacent pixels. Actually, this task is similar to steganalysis, which means to classify the cover image and stego image (adding slight perturbation on cover image for hiding secret message). Moreover, the traditional high-dimension human-design features (SRM) has been verified effective to detect adversarial examples. Recently, neural network based steganalysis methods~\cite{wu2017residual,deng2019fast} perform better than SRM. Drawing insights from these neural network based steganalysis, we design the backbone network shown in Figure \ref{twostream}, which has the following characteristics:
  \begin{itemize}
	  \item Average pooling layer is abandoned in the front layers. Because average pooling layer is a low-pass filter, it reinforces content and suppresses noise-like perturbation signals by averaging adjacent pixels. 
 	  \item The perturbation signal will decay as the layers increases without shortcut connections, resulting in unsatisfying detection performance. As a result, shortcut connections are adopted to preserve the weak perturbation signal.

	  \item  Global covariance pooling (GCP)\cite{DBLP:conf/iccv/LiXWZ17} is introduced for gathering more information. Compared to the first-order statistic (i.e. global average pooling), more useful information can be obtained from the higher-order statistics. 
	  
	  
  \end{itemize}
  
  As for the gradient stream, we adopt the same bone neural network as that of the image stream, resulting from its strong discrimination ability.
  
  \subsubsection{Score Fuser}
  The image and gradient sub-networks are denoted by $F_\text{I},F_\text{G}$, respectively. With the input image as well as the gradient, we can obtain immediate scores $\bm{z}_\text{I}=F_\text{I}(\bm{x})$, $\bm{z}_\text{G}= F_\text{G}(\bm{g})$. Then the immediate scores are mixed to get the final output:
  \begin{equation}
	  \bm{z}' = \bm{z}_\text{I} + \bm{z}_\text{G}
  \end{equation}

  \section{Experiments}
  We now present the experimental results to demonstrate the effectiveness of our method on improving detection performance. 
  \subsection{Setup}
  \subsubsection{Datasets}
  We use two widely studied datasets ImageNet~\cite{ImageNet} and Caltech-256~\cite{Celtach256}. The ImageNet dataset contains 1.2 million training images and the other 50,000 images for testing. Caltech-256 is composed of 256 object categories containing a total of 30,607 images, and we divided it into the training set and testing set by a ratio of 8:2. Here, all images are resized to $224\times224\times3$ color images to match the classification model. 
  \subsubsection{Target Models}
  Different target models are adopted to show the generality of the PACA. For Caltech-256, VGG16~\cite{vgg} is adopted as the classification model. We train the model on the training set using Adam optimizer with learning rate 0.001. For ImageNet, pretrained model ResNet34~\cite{resnet} provided in torchvision is directly adopted. The classification accuracy on the testing set are 78\% for ImageNet, 79\% for Caltech-256. 
  
  \subsubsection{Attack Methods}
  For each target model, we generate adversarial examples from the testing set and use only those that can attack successfully before deploying any countermeasure to the target model in all of our experiments. We conduct untargeted attacks to each target model with five representative
  attack algorithms, SF, EAD, C\&W, DDN, and PGD attacks, as introduced in Section 2. These attacks cover $\ell_1, \ell_2$ and $\ell_\infty$ constraint attacks. We use the default setting for SF, EAD, and DDN.  C\&W under $\ell_2$ constraint is adopted. We use confidence $\kappa=1$ and the number of iterations are 1 and 500, respectively.
  For PGD, $\epsilon=0.03, \alpha = 0.005$ and the number of iterations is 10.
  Our implementations are based on the foolbox and Advertorch.
  
  \subsubsection{Treathen Models}
  \begin{itemize}
	  \item \emph{Oblivious adversaries} have a full access and knowledge to classifier $F$ but are not aware of detector $D$ in place.
	  \item \emph{Omniscient adversaries} know the model details of both classifier $F$ and detector $D$. 
  \end{itemize}
  
  \subsubsection{Training Details of PACA}
  The Adamax optimizer was used with a mini-batch of 32 shuffled clean and adversarial images. The batch normalization parameters were learned via an exponential
  moving average with decay rate 0.05. For the fully-connected classifier layer, we initialized the weights with xavier uniform distribution and no bias. 
  The learning rate is 0.001 and dropped by a factor of 0.1 at 30, 70 and 150 epochs,with a total budget of 200 epochs.
  All the experiments are implemented by PyTorch, and the codes will be released later. For quick convergence, we suggest initializing the model with the parameters of the pretrained model which detects large-perturbation adversarial examples.
  
  
  \begin{figure}[t]
	  \centering 
	  \subfigure[PACA-ImageNet]{ 
		\label{fig:subfig:a} 
		\includegraphics[width=1.5in]{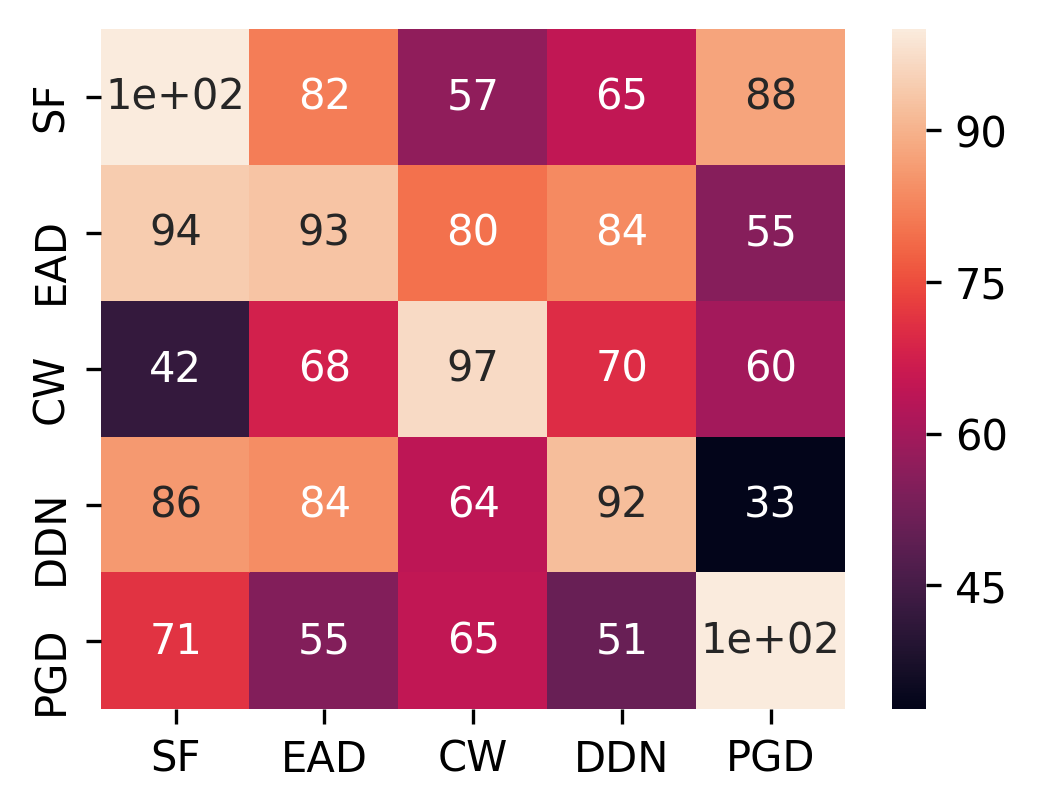}} 
	  \subfigure[SRM-ImageNet]{ 
		\label{fig:subfig:b} 
		\includegraphics[width=1.5in]{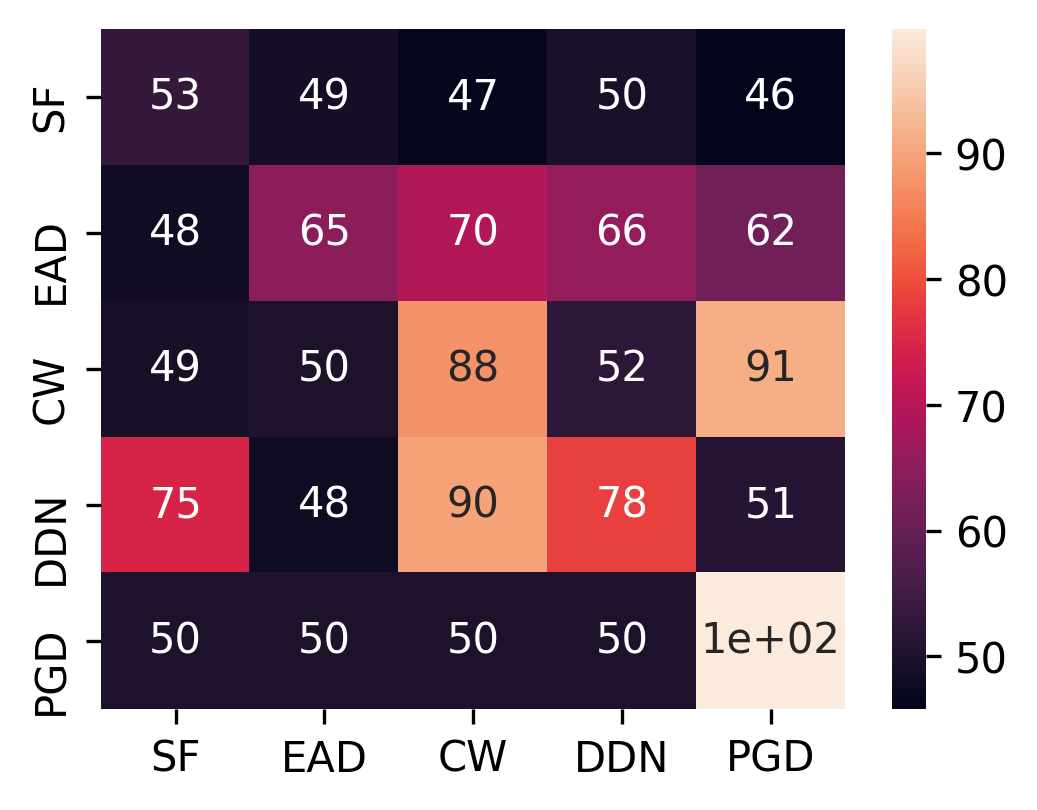}}\\\vspace{-1em}
	  \subfigure[PACA-Caltech-256]{ 
		  \label{fig:subfig:b} 
		  \includegraphics[width=1.5in]{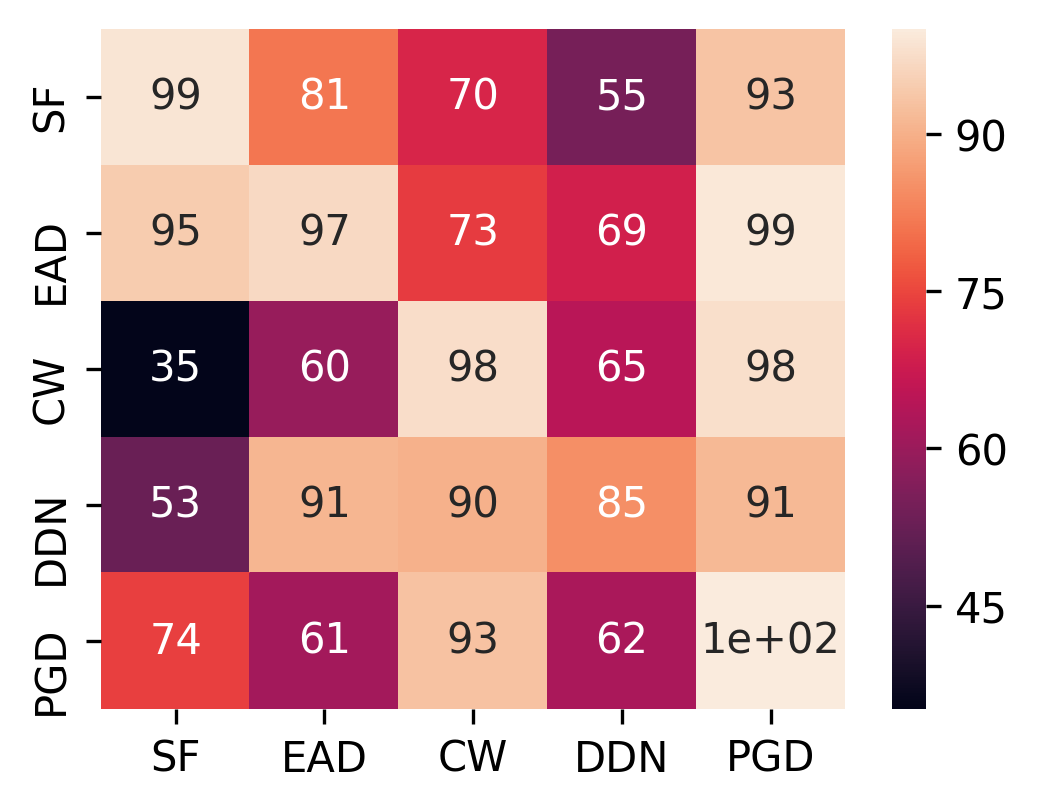}} 
	  \subfigure[SRM-Caltech-256]{ 
		  \label{fig:subfig:b} 
		  \includegraphics[width=1.5in]{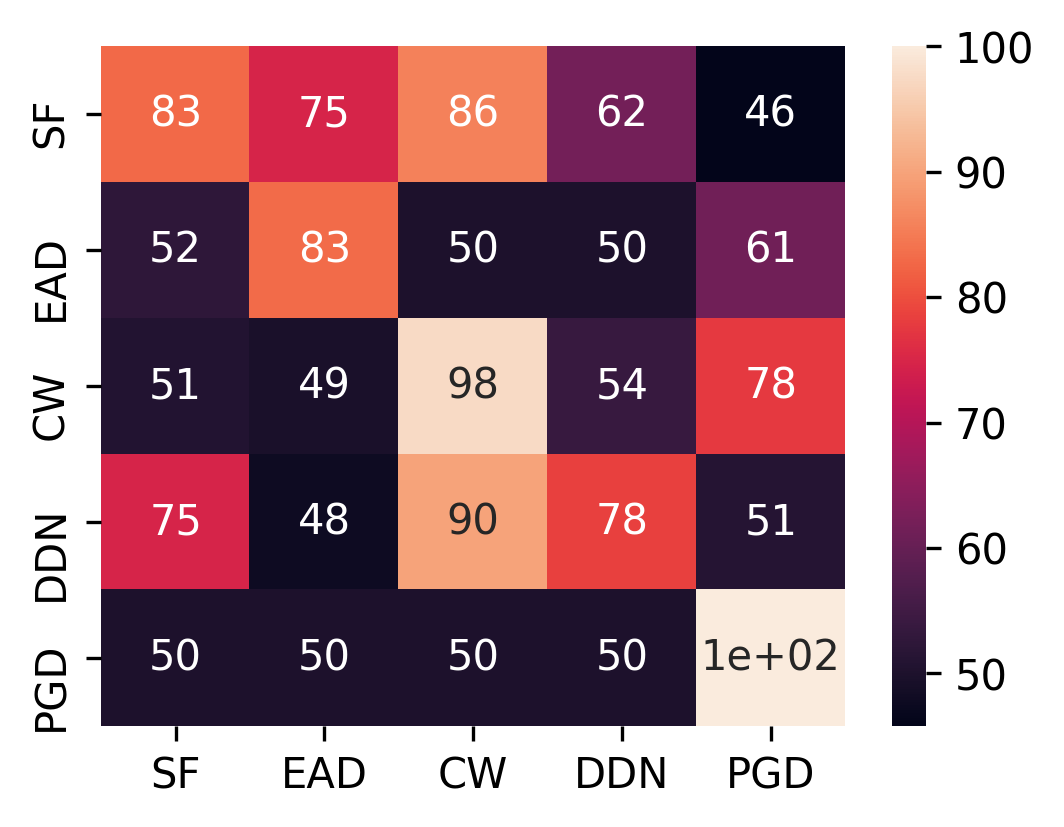}} \\\vspace{-1em}
	  \caption{The generalizability detection accuracy (\%) of PACA and SRM on two datasets. The detection accuracy of PACA is higher than SRM in most cases, meaning that the transferability of PACA is better than SRM.} 
	  \label{transfer} 
	  \vspace{-1em}
  \end{figure}

  \subsection{Performance under the Oblivious Adversary}
  Table \ref{result_imagenet} and Table \ref{result_Caltech} give the detection performance under the oblivious attack. The detection methods FS, LID, MDA, SRM are adopted for comparison. PGD is easiest to be detected, and the detection accuracy of all detection methods is nearly 100\%. EAD is one of the most difficult methods to be detected, for its perturbation is slight as well as its confidence is not quite low. For the methods except PGD, the proposed method PACA outperforms other methods on two datasets with a clear margin. The advantage of detection accuracy even approaches 20\% with respect to all methods when detecting DDN. These results verify the effectiveness of PACA under the oblivious adversary.
  
  We have also tested the generalization detection performance, since in most cases the detector has no knowledge of which algorithm the adversary adopted. The generalization detection experiments show the generalizability of detection methods among different attacks.
  Figure \ref{transfer} shows generalizability heatmaps by PACA comparing with SRM on two datasets. The detectors are trained with one of the attacks listed in the columns and tested against one another listed in the rows. For SRM, the transfer performance is unsatisfying, since many results are around 50\% on both datasets. Compared with SRM, PACA owns far better generalizability. The PACA detector trained on SF or EAD can detect  other attacks effectively on two datasets. Analyzing these two attacks, we find that they are both under $\ell_1$ constraint. Similarly, the detector trained on DDN and C\&W has ability to detect other methods, for they are under $\ell_2$ constraint.

  \begin{table}[t]
	\small
	  \centering
		  \caption{Detection accuracy (\%) under the Oblivious Adversary on ImageNet. }
		  \scalebox{0.9}{
	  \begin{tabular}{c|ccccc}
	  \toprule
		& FS &LID& MDA & SRM & PACA \\
	  \midrule
	  SF  &  61.30 &  59.67 &66.38 &  53.39 &\textbf{98.30}\\
	  EAD &  54.75 &  60.88 &68.70 &  74.82 & \textbf{89.22} \\
	  C\&W  &  55.47 &  64.25 &68.87 &  87.24 & \textbf{96.05} \\
	  DDN  &  69.63 &  61.41 &68.52 &  65.62  & \textbf{91.47} \\  
	  PGD  &  95.55 &  99.21 &99.55 &  \textbf{99.68}  & 99.32 \\
	  \bottomrule
	  \end{tabular}}
	  \vspace{-1em}
	  \label{result_imagenet}
  \end{table}
  \begin{table}[t]

	\small
	  \centering
	  \caption{Detection accuracy (\%) under the Oblivious Adversary on Caltech-256. }
	  \scalebox{0.9}{
	  \begin{tabular}{c|ccccc}
	  \toprule
		& FS &LID& MDA & SRM & PACA \\
	  \midrule
	  SF  &  59.97 &  73.94 &80.61 &  82.82 &\textbf{97.19}  \\  
	  EAD &  52.07 &  71.42 & 74.08 &  83.48 & \textbf{90.71}  \\ 
	  C\&W  &  58.25 &  70.29 & 73.24 &  97.59 & \textbf{97.97} \\
	  DDN  &  61.17 &  69.49 &74.19 &  78.10  & \textbf{97.29} \\
	  PGD  &  100 &  97.13 &99.97 &  \textbf{99.78}  & 99.32 \\
	  \bottomrule
	  \end{tabular}}
	  \vspace{-1em}
	  \label{result_Caltech}
  \end{table}

  \subsection{Performance under the Omniscient Adversary}
  When the adversaries have full knowledge of the classifier as well as the detector, they can generate adversarial examples deceiving both.
  Actually, this attack is also named second-round attack, which has been used to evaluate the performance of detectors in  ~\cite{DBLP:conf/ccs/Carlini017,DBLP:conf/nips/PangDDZ18}. Following ~\cite{DBLP:conf/nips/PangDDZ18}'s setting, here we evaluate the performance of the proposed scheme. Adopt C\&W as the original attack, and then modify the C\&W attack by introducing to its adversarial objective an additional loss term for penalizing being detected:
  \begin{equation}
	  \min \left\{\|\boldsymbol{\delta}\|_{2}+J_{F}(\bm{x}+\boldsymbol{\delta})+ J_{D}(\bm{x}+\boldsymbol{\delta})\right\}
  \end{equation}
  where $J_F$, $J_D$ are the loss of classifier $F$ and proposed detector $D$, respectively.
  The modified version is named C\&W-PACA. Attack successful rate and average $\ell_2$ distance between adversarial and clean images are utilized to measure the defense ability. 
  Large $\ell_2$ distance means that it is more hard to generate adversarial examples.
  Table \ref{CWL2} shows that the successful rate of C\&W-PACA is far smaller than that of C\&W, and the $\ell_2$ distance of modified attack is higher in both datasets, indicating that PACA does enhance the defense ability.
  \begin{table}[t]
	\small
	  \centering
		  \caption{The attack successful rate and the average $\ell_2$ distance of C\&W and C\&W-PACA on ImageNet and Caltech-256. }
		  \scalebox{0.9}{
	  \begin{tabular}{c|ccc}
	  \toprule
	& Attacks   &Successful rate& $\ell_2$ distance  \\
	  \midrule
	ImageNet &   C\&W  &  77.80\% &  0.1430 \\
	 & C\&W-PACA  &  7.00\% & 0.1594   \\
	 \hline
	 Caltech-256 & C\&W  &  77.50\% &  0.1463 \\
	  &C\&W-PACA  &  8.90\% & 0.1600   \\
	  \bottomrule
	  \end{tabular}}
	  \label{CWL2}
  \end{table}
  
  \subsection{Ablation Study}
  To inspect the effect of each component of PACA, we conduct the control experiments on ImageNet by removing or replacing the component. DDN and C\&W are chosen as the attack methods, which represent different perturbation attacks. The results are shown in Table \ref{ablationstudy}. PACA performs best among different settings. Gradient stream performs well on detecting DDN, while the image stream does better in detecting C\&W. That is to say, two steams are complementary for they can cope with different attacks. Besides, replacing the GCP or removing shortcut connection will destroy the detection performance, meaning that these architectures play positive roles in PACA.  
  \begin{table}[t]
	\small
	  \centering
		  \caption{Detection accuracy (\%) of variant detectors of PACA.}
		  \scalebox{0.9}{
	  \begin{tabular}{cccccc}
	  \toprule
	  Operations  &DDN&  C\&W  \\
	  \midrule
	  PACA  & \textbf{91.47}  & \textbf{96.05} \\
	  Remove image stream  &  91.07 & 68.47  \\
	  Remove gradient stream  &  76.19 & 92.16 \\
	  GCP$\rightarrow$GAP &  89.36 &94.23  \\
	  Remove short-cut connection& 89.04  & 66.35  \\
	  Single logits + FC & 75.75 & 60.61 \\
	  \bottomrule
	  \end{tabular}}
  
	  \label{ablationstudy}
  \end{table}
  Moreover, we also investigate the performance of directly using prediction logit for classification rather than using the gradient. Three fully-connected layers and ReLU activation are adopted for classification, and the number of neurons are 512, 32, respectively. The results in the last row in Table \ref{ablationstudy} show that only using logits assembled by fully-connected layers for classification is undesirable, the detection accuracy is lower than that of merely using gradient stream (Remove image stream), indicating that the gradient stream does favor to detection, which exploits the information of the classification model.
  

  \section{Conclusion}
  The newly proposed methods like DDN, EAD which only require slight perturbation are hard to detect by existing detection methods. Through exploring the perturbation and the predicted logits of these adversarial examples, we find there exists compliance between perturbation and prediction confidence. Slight perturbation leads to low prediction confidence. For fully exploiting the confidence information as well as the classification model, the gradient is introduced by back-propagating the confidence loss to the images, where the confidence loss is defined as the cross-entropy between prediction logits and its one-hot vector. To make use of the information of image artifacts as well, we propose a two-stream convolutional neural network for detecting different types of attacks, including image stream and gradient stream. Extensive experiments have been performed to evaluate the performance of the proposed PACA, and the results show that PACA owns stronger detection ability as well as better generalizability in most cases.
  
\clearpage
\bibliographystyle{IEEEbib}
\bibliography{ijcai20}

\end{document}